\documentclass[conference]{IEEEtran}
\IEEEoverridecommandlockouts
\usepackage{cite}
\usepackage{amsmath,amssymb,amsfonts}
\IEEEoverridecommandlockouts
\usepackage{algorithmic}
\usepackage{algorithm}
\usepackage{graphicx}
\usepackage{textcomp}
\usepackage{xcolor}
\usepackage{import}
\usepackage{soul}
\usepackage{hyperref}
\usepackage{environ}
\usepackage{mathrsfs}
\usepackage{morewrites}
\usepackage{hyperref} 
\usepackage{enumitem}
\usepackage{lettrine}
\usepackage{tikz}
\usetikzlibrary{shapes.geometric, arrows}
\usepackage{amsmath}
\usepackage{caption}
\usepackage{transparent}
\usepackage{booktabs} 
\usepackage{siunitx}   
\usepackage{multirow}

\def\BibTeX{{\rm B\kern-.05em{\sc i\kern-.025em b}\kern-.08em
    T\kern-.1667em\lower.7ex\hbox{E}\kern-.125emX}}
\begin{document}

\title{Fed-MUnet: Multi-modal Federated Unet for Brain Tumor Segmentation}

\author{
    \IEEEauthorblockN{
        Ruojun Zhou\IEEEauthorrefmark{1}, 
        Lisha Qu\IEEEauthorrefmark{1}, 
        Lei Zhang\IEEEauthorrefmark{1}, 
        Ziming Li\IEEEauthorrefmark{2}, 
        Hongwei Yu\IEEEauthorrefmark{2},
        Bing Luo\IEEEauthorrefmark{1}
    }
    \IEEEauthorblockA{\IEEEauthorrefmark{1}Duke Kunshan University, Kunshan, China}
    \IEEEauthorblockA{\IEEEauthorrefmark{2}Wuhan University, Wuhan, China}

        \thanks{This work is supported by Suzhou Frontier Science and Technology Program (Project SYG202310), Kunshan Municipal Government research funding (Project 24KKSGR013), and DKU Foundation Hou Tu Research (HTR) Funding. Corresponding author: Bing Luo (bing.luo@dukekunshan.edu.cn).}
}

\maketitle

\begin{abstract}
Deep learning-based techniques have been widely utilized for brain tumor segmentation using both single and \mbox{multi-modal} Magnetic Resonance Imaging (MRI) images. Most current studies focus on centralized training due to the intrinsic \mbox{challenge} of data sharing across clinics. To mitigate \mbox{privacy} \mbox{concerns}, researchers have introduced Federated \mbox{Learning} (FL) methods to brain \mbox{tumor} \mbox{segmentation} tasks. However, \mbox{currently} such methods are \mbox{focusing} on single modal MRI, with \mbox{limited} study on multi-modal MRI. The challenges \mbox{include} \mbox{complex} \mbox{structure}, large-scale parameters, and overfitting \mbox{issues} of the FL based methods using multi-modal MRI. To address the above challenges, we propose a novel multi-modal FL framework for brain tumor segmentation (Fed-MUnet) that is suitable for FL training. We evaluate our approach with the BraTS2022 datasets, which are publicly available. The \mbox{experimental} \mbox{results} \mbox{demonstrate} that our \mbox{framework} achieves FL nature of \mbox{distributed} learning and privacy preserving. For the enhancing tumor, tumor core and whole tumor, the mean of five major metrics were 87.5\%, 90.6\% and 92.2\%, \mbox{respectively}, which were higher than SOTA methods while preserving privacy. In terms of parameters count, quantity of floating-point operations (FLOPs) and inference, Fed-MUnet is Pareto optimal compared with the state-of-the-art segmentation backbone while achieves higher performance and tackles privacy issue. Our codes are open-sourced at \href{https://github.com/Arnold-Jun/Fed-MUnet}{https://github.com/Arnold-Jun/Fed-MUnet}. 
\end{abstract}


\section{\textbf{Introduction}}
\lettrine{S}{egmenting} brain tumors from multi-modal Magnetic Resonance Imaging \mbox{(MRI)} images is clinically important for \mbox{cancer} diagnosis and treatment. However, the \mbox{variability} in \mbox{tumor} location, shape, and appearance complicates precise \mbox{segmentation}. The advent of deep learning (DL) has led to the proposal of numerous advanced technologies for brain tumor segmentation. Since MRI imaging can be multi-modal by \mbox{nature} if acquired using different imaging \mbox{sequences}, such as T1-w, T1c, T2-w, and FLAIR, an increasing amount of researchers focus on modern architecture to deal with \mbox{multi-modal} input, as exemplified by \cite{A30, A17, A7, A8}. The great \mbox{success} of these multi-modal methods relies highly on large-scale and centralized datasets.

However, it is typically challenging to directly gather and exchange medical images across hospitals or organizations in real-world practice. Strict privacy protection regulations, including the Health Insurance Portability and \mbox{Accountability} Act (HIPAA) \cite{A10} and the General Data Protection \mbox{Regulation} (GDPR) \cite{A11}, make it difficult to share medical imaging data across hospitals. For instance, the European Union (EU) GDPR clearly stipulates that the transfer of personal data is limited to the EU or countries/organizations. To \mbox{address} \mbox{privacy} issues, Luo \cite{f2, f3} extended the findings of cost-effective FL framework. In FL, the raw data is kept on the \mbox{local} device and not uploaded to the central server or \mbox{elsewhere}. Only intermediate results, such as model \mbox{parameters} or \mbox{gradient} updates, are encrypted and transmitted, which cannot be directly translated to the original data, thus \mbox{protecting} data privacy. Recent works \cite{A12, A14, A15} implemented FL in medical tasks, aiming to address the privacy issue in brain tumor \mbox{segmentation}. However, the methods in \cite{A12, A14} perform tumor segmentation using the single MRI as input due to \mbox{complexity} of multi-modal images. The method in \cite{A15} \mbox{directly} uses U-Net\cite{A9} for segmentation task on multi-modal images without further exploration of modern multimodal models suitable for federated learning.

 Consequently, there is a lack of study on the application of FL to multi-modal MRI brain tumor segmentation tasks. There are two major challenges in introducing FL to multi-modal brain tumor segmentation model: 
 
$\bullet$ First, some multi-modal models have a two-stage \mbox{structure}, which greatly reduces the efficiency of FL \mbox{training}. For instance, recent work \cite{A8} proposed pre-training a \mbox{variational} auto-encoder (VAE) and used the pre-trained VAE in the segmentation backbone. When combined with FL, this type of two-stage model increases the \mbox{communication} \mbox{overhead} and increases the risk of training interruption due to \mbox{network} anomalies or equipment failures. From this \mbox{perspective}, a one-stage end-to-end architecture is preferred in the FL setting. 

$\bullet$ Second, the existing multi-modal segmentation models have a large number of parameters for dealing with multi-modal input, which are not conducive to FL training. In FL, \mbox{overfitting} is a common problem. The large number of parameters in multi-modality models such as DeepLabv3+  \cite{A17} exacerbates overfitting. 


To address the above challenges, in this paper, we propose a novel transformer-based FL framework (\mbox{Fed-MUnet}) for multi-modal brain tumor segmentation. Our model follows client-server architecture, and clients train the local dataset using a newly proposed segmentation backbone, Multi-modal U-net (M-Unet). M-Unet leverages the advantages of U-Net, such as efficient structure and fewer parameters, while having the capability of multi-modal feature fusion. To mitigate the problem of overfitting in FL, we design a lightweight Cross Modality Module (CMM) that helps the model achieve SOTA segmentation accuracy without significantly increasing the number of model parameters in the FL paradigm. CMM takes advantage of the powerful ability of transformer to capture global information, which effectively integrates multi-modal image features. Some advanced deep learning models have been proposed in multi-modal brain tumor segmentation, such as introVAE \cite{A8} and Deeplabv3+ \cite{A17}. However, most of these models do not fit the FL framework. We summarize the key contributions of work as follows: 
\begin{itemize}[leftmargin=2em]  
    \item We develop an end-to-end segmentation model \textbf{M-Unet} incorporating a \textbf{CMM} that effectively addresses the low training efficiency issue and alleviates overfitting in FL. To the best of our knowledge, we are the first to integrate FL with \mbox{multi-modality} in brain tumor \mbox{segmentation}. With the lightweight model, we \mbox{improve} the \mbox{performance} of segmentation \mbox{accuracy} \mbox{without} \mbox{increasing} the \mbox{complexity} and number of \mbox{parameters}.
    \item We propose a novel FL framework \mbox{\textbf{(Fed-MUnet)} }for multi-modal brain tumor segmentation, where differential privacy is implemented to maintain the \mbox{confidentiality} of patient data. This approach balances high segmentation performance with privacy protection.
    \item The evaluation results on BraTS2022 show that our framework achieved comparable accuracy to \mbox{centralized} learning methods: for the enhancing tumor, tumor core and whole tumor, the mean of five major metrics were \textbf{(87.5\%)}, \textbf{(90.6\%)} and \textbf{(92.2\%)}, respectively, which were higher than SOTA methods. 

\end{itemize}

\section{\textbf{Preliminary}}
In this section, we first present basic information about brain tumor segmentation. In the second part, we detail the FL setting for our model deployment.

\subsection{\textbf{Background}}
Brain tumor segmentation is a very important part in clinical medicine. In order to better diagnose brain \mbox{tumors}, MRI data sets are widely used in the diagnosis and \mbox{treatment} of brain \mbox{tumor} diseases in hospitals. MRI has the \mbox{characteristics} of multi-parameter and multi-sequence \mbox{imaging}, which can \mbox{reflect} different information of the \mbox{tumor} area and \mbox{provide} a rich data basis for brain \mbox{tumor} \mbox{segmentation}. Our \mbox{research} aims to develop a privacy-preserving high-precision \mbox{multi-modal} FL framework to process MRI datasets for \mbox{clinical} diagnosis.

\subsection{\textbf{Federated Learning Setting}}
Let $\mathcal{D}$ denotes the MRI dataset and each dataset $X_K$ for client $K$ is random sampled from $\mathcal{D}$. Let $w_K$ represents the proportion of the entire dataset $\mathcal{D}$ for $X_K$ assigned to the K-th client, $w = \{(w_1, w_2, \ldots, w_N) \mid \! w_i \! \geq \! 0, \sum_{i=1}^N w_i = 1, \text{for} \ i = 1, 2, ..., N \}$. The vector $w$ is a concrete sample drawn from dirichlet distribution, that is, $w = (w_1, w_2, \ldots, w_N) \sim Dir(w \! \mid \! \alpha)$.\footnote{$Dir(w\mid\alpha) = \frac{1}{B(\alpha)} \prod_{i=1}^N w_i^{\alpha_i - 1}$, where $B(\alpha)$ is the multivariate Beta function and hyperparameter $\alpha$ controls the variance of dirichlet distribution.} For total modality $\mathcal{M}$, local dataset$X_{K}$ = \{$X_{K}^1, X_{K}^2, ...., X_{K}^{\mathcal{M}}$\}.
 
\begin{figure}
\centering
\includegraphics[height=5cm,width=9cm]{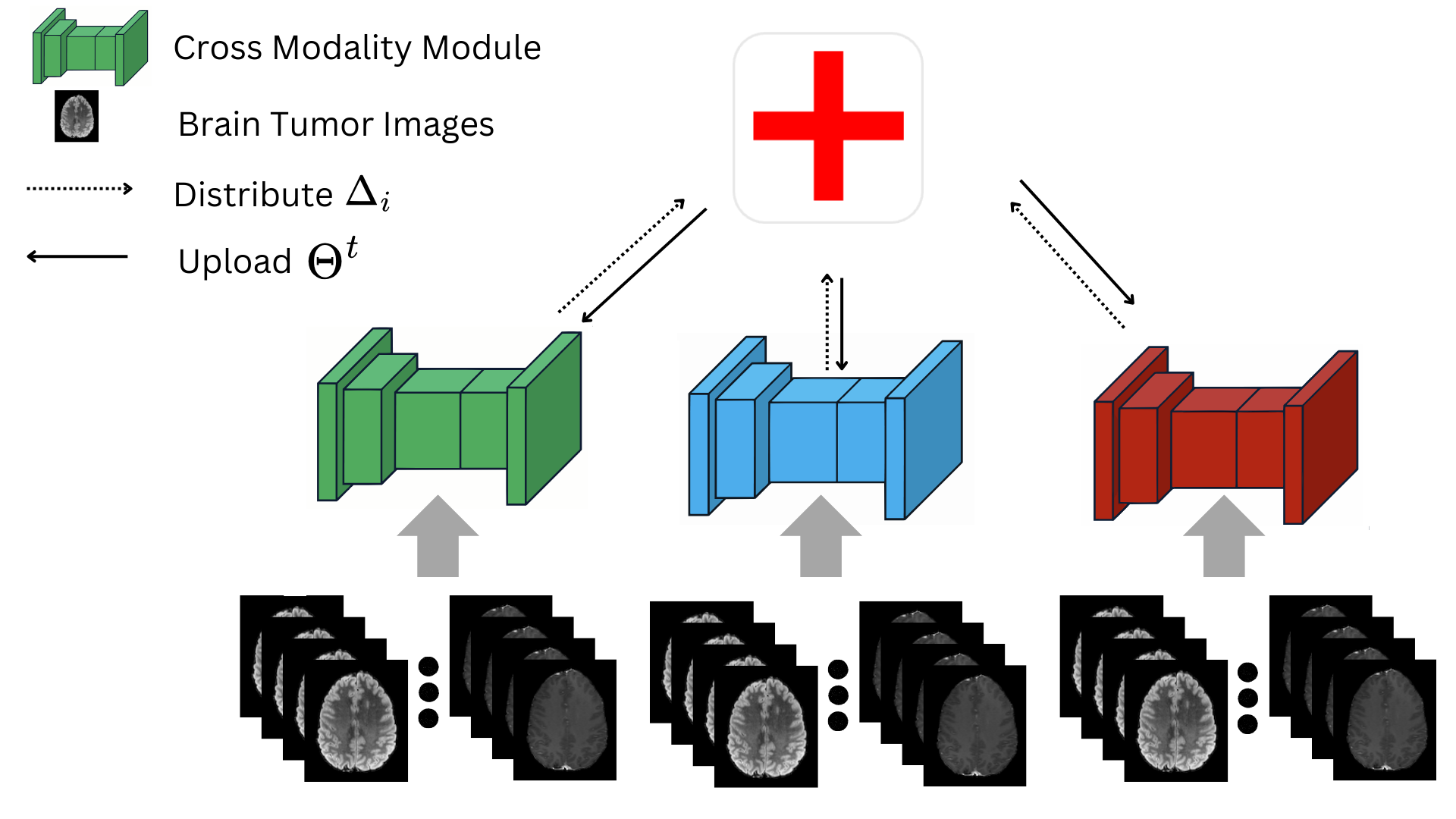}
\caption{ Decentralized Training Paradigm for Brain Tumor Segmentation.}
\label{1}
\end{figure}

Fig. \ref{1} illustrates the decentralized training paradigm of our model, which adheres to a typical client-server structure. Each hospital performs forward propagation using local datasets $X_K$ and local model $\theta_K$. After transferring the computed parameter differential $\Delta_K$ back to the central server for global aggregation, the updated model $\Theta^t$ is then sent to the local client for next global training epoch.

\section{\textbf{Method}}

Aiming to achieve privacy preservation, we adopt \mbox{differential} privacy (DP) as a powerful privacy protection mechanism for secure transmission and aggregation of model parameters. In order to tackle the challenge that multi-modal models are difficult to deploy in the FL, we propose a novel segmentation model Multi-modal U-net (M-Unet), which is primarily based on transformer and U-net.

In this section, we illustrate FL algorithm our local \mbox{segmentation} model M-Unet in detail. High performance achieved by our model Fed-MUnet can be attributed to M-Unet. In the first part, we elaborate on the specific algorithms for updating and transmitting model parameters. Then, we explain the original intention of the structure design. In the third part, we describe the operation mechanism and design reason of CMM.

\subsection{\textbf{Federated Learning with Differential Privacy}}
In our work, we utilize DP-FedAvg \cite{A19} to achieve privacy protection. The specific procedures are detailed in Algorithm \ref{alg:client} and Algorithm \ref{alg:server}. 

\subsubsection{\textbf{Client Side}}
\begin{algorithm}
\footnotesize
\caption{\textbf{Client Side}}
\label{alg:client}
\begin{algorithmic}[1]
\STATE \textbf{Input:} Global model: $\Theta^t$, where t is the global epoch; loss function $l(\theta)$; hyperparameters: learning rate $\eta$; local epochs $e$; batch size: $B$; local model: $\theta$;  Parameters upload: $\Delta_K = \theta - \Theta^t$.
\STATE Initialize local model $\theta \leftarrow \Theta_0$
\FOR{each local epoch $j = 1, 2, \ldots, E$}
    \STATE Split local data into different batches $\mathcal{B}_k$
    \FOR{each batch $b \in \mathcal{B}_k$}
        \STATE Perform gradient descent: $\theta \leftarrow \theta - \eta \nabla l(\theta; b)$
        \STATE Clip model parameters: $\theta \leftarrow \Theta^t + FlatClip(\theta - \Theta^t)$  \quad $\triangleright$  Eq.( \ref{eq:flatclip})
    \ENDFOR
\ENDFOR \\
\end{algorithmic}
\end{algorithm}
On the client side, initialized\textit{/}aggregated parameters $\Theta^t$ are transferred from server for local training. Notably, the clipping operation $Clip(\theta, \Theta^t)$ ensures that there are no extreme values or outliers that could significantly skew the results. Additionally, clipping restricts the maximum sensitivity of the model’s output with respect to a single data point, thereby allowing less noise to be added while still maintaining the desired level of privacy:
\begin{equation}
\small
\label{eq:flatclip}
    FlatClip(\Delta_K) = \Delta_K \cdot \min \left( 1, \frac{S}{\|\Delta_K\|} \right),
\end{equation}
where S is the prior upper bound of $\|\Delta_K\|$.

\begin{algorithm}
\footnotesize 
\caption{\textbf{Server Side}}
\label{alg:server}
\begin{algorithmic}[1]
\STATE \textbf{Input:} Learning rate $\eta_t$, gradient clipping bound $C$; noise parameter $\sigma$; batch size for each round $B$; number of  training examples generated by client k $n_K$; per-user example cap $\hat{w}$.
\STATE Randomly initialize model parameter $\Theta^0$
\STATE Define client weight: $w_k$ for client $k$ as $w_k = \min(\frac{n_k}{\hat{w}}, 1)$
\STATE Set $W = \sum_k w_k$
\FOR {each round of global epoch $t = 1, 2, \ldots, T$}
    \STATE Select a sample group of clients $\mathcal{C}^t$ for the current iteration with probability $q$
    \FOR {each client $k \in \mathcal{C}^t$}
        \STATE Execute local training: $\Delta_K = \theta - \Theta^{t-1}$
    \ENDFOR
    \STATE Aggregate client updates: \quad$\Delta^t = Agg(q, W, \Delta_K)$ \quad\quad\quad $\triangleright$  Eq. (\ref{eq:agg})
    \hspace{-5pt}
    \STATE Clip operation: \quad$\Delta^t = Clip( \Delta^t , C)$  \quad\quad\quad\quad\quad\quad\quad\quad\quad $\triangleright$  Eq. (\ref{eq:clip})

    \STATE \text{Update global model parameter}:
$
    \centering{\Theta_t \leftarrow Update(\Theta_{t-1}, \Delta^t, \sigma)}     \quad\quad\quad\quad\quad\quad\quad\quad\quad\quad\quad\quad \ \triangleright  \text{Eq. (}\ref{eq:update}\text{)}
$
\ENDFOR \\
\end{algorithmic}
\end{algorithm}

Notably, clients results will be randomly sampled with probability $q$ for global updating to further preserve privacy. For efficiency, in each global epoch, $q \times 10^2$ percent of clients will perform local training. 

\begin{figure*}
\vspace{-40pt}
\centering
\includegraphics[height=8.5cm,width=18cm]{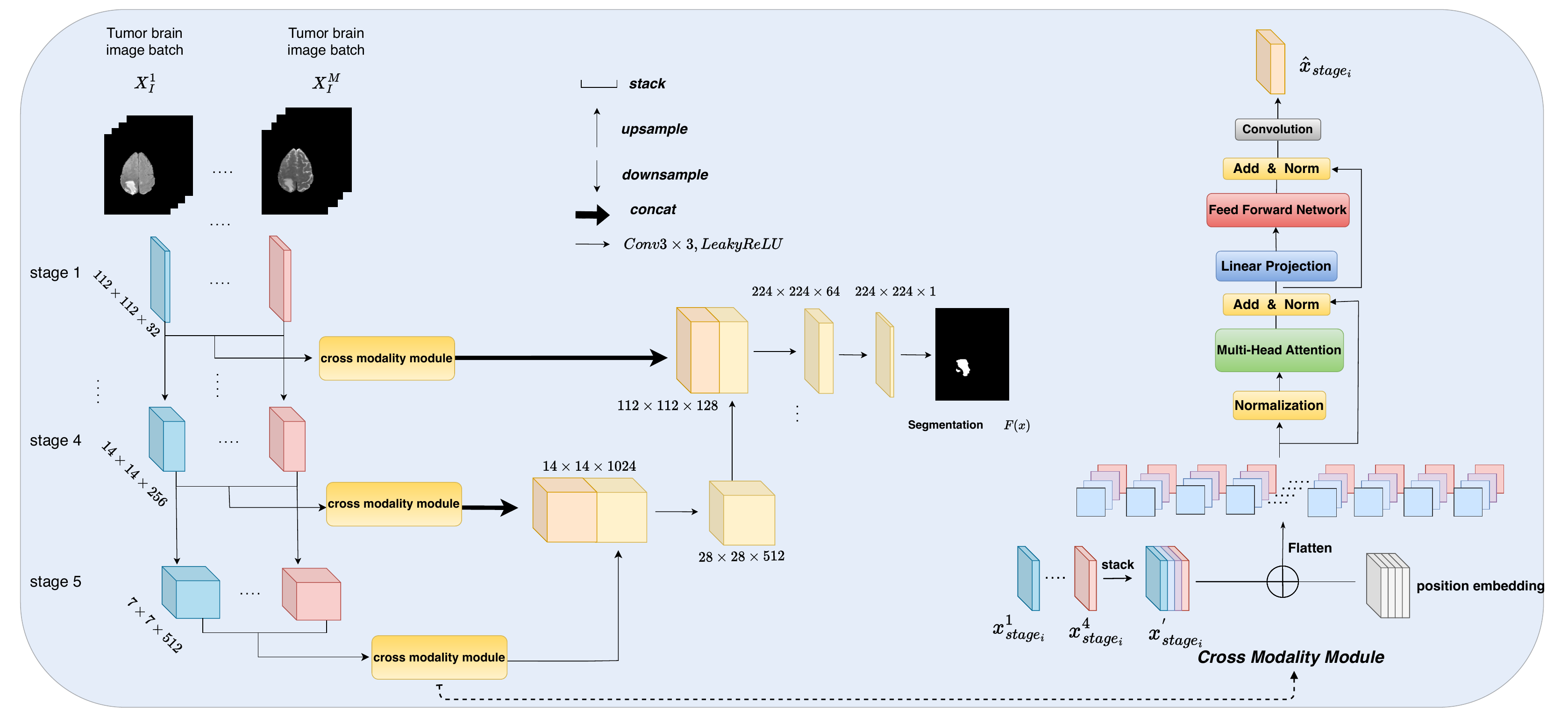}
\caption{The framework of our segmentation backbone M-Unet. The right part is the structure of CMM. The overall framework follows the encoder-decoder architecture of Unet and a Transformer-like module CMM is used for multi-modal feature integration. For concatenation, we deploy a convolution operation to align the dimension of upsampling feature matrix.}
\label{2}
\end{figure*} 
\subsubsection{\textbf{Server Side}}
On the server side, the parameter \mbox{differential} collected from all clients is aggregated via :
\begin{equation}
\small
\label{eq:agg}
    Agg(q, W, \Delta_K) = \frac{\sum_{k \in C^t} w_k \Delta_K}{qW}.
\end{equation}
To prevent training instability caused by gradient explosion, we also adopt Clip operation on server side:
\begin{equation}
\small
\label{eq:clip}
    Clip (C, \Delta^t) = \Delta^t / \max(1, \frac{\|\Delta^t\|}{C}).
\end{equation}
Finally, Gaussian noise and $\Delta^t$ are used to update the global model parameters. The variance $\sigma$ is a hyperparameter to manually control the strength of privacy protection:
\begin{equation}
\label{eq:update}
\small
    Update(\Theta_{t-1}, \Delta^t, \sigma) = \Theta_{t-1} + \Delta^t + N(0, I\sigma^2).
\end{equation}

\subsection{\textbf{The Segmentation Backbone}}
 We naturally adopt U-net due to its outstanding \mbox{performance} in medical segmentation tasks. References \cite{A9} and \cite{A20} demonstrate several advantages of U-net, including its \mbox{symmetric} \mbox{encoder-decoder} architecture, skip connections and properties suitable for few-shot learning. As shown in Fig. \ref{2}, multi-modal images are first fed into a convolutional neural network (CNN) to extract different levels of \mbox{features}. Due to the locality and weight sharing characteristics of \mbox{convolution} operation, CNN has high training efficiency when dealing with large-scale image data. To fit into the FL setting, we chose CNN as the feature extractor for \mbox{downsampling} instead of Transformer to reduce parameters and model \mbox{complexity}. Apart from the symmetric end-to-end structure, we also \mbox{propose} a CMM for processing \mbox{multi-modal} medical \mbox{features}. The attention-based CMM aims to integrate \mbox{multi-modal} \mbox{features} obtained from CNN for enhanced feature upsampling. 
\subsection{\textbf{Cross Modality Module}}
CMM is mainly based on Transformer. Through the self-attention mechanism, Transformer model can directly calculate the dependency between any two locations, so as to effectively capture long-distance dependencies. Traditional CNN models, however, can only capture local information when dealing with multi-channel image. Therefore, we use Transformer-like structure in CMM to better integrate multi-modal features.
As shown on the right side of Fig. \ref{2}, we start with features of four modalities, each of size $C \times H \times W$. To obtain cross-modality feature matrix, we perform the following steps:

\begin{enumerate}
    \item \textbf{Stack the Images:} First, we stack the features $x_{stage_i}^j$ along the channel dimension to form a single tensor of size $4C_i \times H_i \times W_i$.  Position embedding is added to the stacked features to help the model understand the sequential relationship between elements and patches.
    \item \textbf{Flatten the Tensor:} Next, we flatten the tensor into a 2D matrix. Specifically, we reshape the tensor from $(4C_i \times H_i \times W_i)$ to $4C_i \times (H_i \cdot W_i$).
\item \textbf{CMM Output:} The resulting matrix $\hat{x}_{stage_i}$ is used as the input to the transformer. In the last phrase of CMM, a $Conv2d$ is used to map the fused features $\hat{x}_{stage_i}$to align the dimension of upsampling matrix $ \in \mathbb{R}^{4C_i \times \frac{H_i}{2} \times \frac{W_i}{2}}$ .
\end{enumerate}
The loss function of our method contains two parts.  The Dice loss is defined by:
\begin{equation}
    \small 
    \mathcal{L}_{Dice} =   -\frac{2 \sum_{x \in \Omega} F(x) \cdot Y(x)}{\sum_{x \in \Omega} F(x)^2 + \sum_{x \in \Omega} Y(x)^2} .
\end{equation}
Dice loss focuses on the global segmentation effect. To make the model focus on the classification accuracy of each pixel, we added pixel-wise cross entropy loss:
\begin{equation}
    \small
    \mathcal{L}_{CE}= -\frac{\sum_{x \in \Omega} \left[ Y(x) \log(F(x)\! +\! (1\! - \!Y(x) \log(1\! - \!F(x)) \right]}{\sum_{x \in \Omega}1} ,
\end{equation}

where the x is the pixel of image area $\Omega$, $F(x)$ is the model output and $Y(x)$ is the ground truth.
\vspace{-20pt}
\section{\textbf{Experiments}}

We train and test our model using the public dataset Brats2022. We compare our \mbox{method} with other \mbox{methods} on six metrics: Dice score, Jaccard index, Hausdorff \mbox{distance}, \mbox{Sensitivity}, Precision and Specificity. We also \mbox{valuate} the \mbox{complexity} of our model compared with other high-performance models. 

\subsection{\textbf{Experimental Settings}}
\subsubsection{\textbf{Datasets}}
 The BraTS22-GLI challenge dataset\footnote{Since the test data of the BraTS challenge is not accessible to the public, we mix the accessible training and validation sets and randomly split all samples into 70\%, 20\% and 10\% for training, validation and testing, respectively.} is used, including multiparametric MRI (mpMRI) scans of brain \mbox{tumors} from routine clinical acquisitions at different centers as test, validation, and training data. The Brats dataset is a public medical image dataset provided by multiple medical centers for the research and development of brain tumor segmentation algorithms. The training and validation data are the same as those used in BraTS21.  Expert neuroradiologists identify the genuine annotations of the tumor subregions. All BraTS mpMRI scans are provided as NIfTI files (.nii.gz). 

A total of xyz patients are included in this study. Each patient contains MRI Images of four modalities, each with a dimension of $240\times240\times155$. The four modalities, as shown in Fig. \ref{3} have different properties and different roles in clinical diagnosis: T1 imaging for anatomical structure; T1c for tumor enhancement by injecting contrast agent into the blood of the subject before imaging. T2 imaging for higher lesion visibility; FLAIR is a T2 hydrographic image (suppressing the high signal of the cerebrospinal fluid), which is brighter when the water content is large and can determine the peritumoral edema area. 

\vspace{-20pt}
\hspace{-100pt}
\begin{table*}[t]
\vspace{-15pt}
\caption{Evaluation Results of Our Method Compared with Other SOTA Frameworks.
\centering{Our proposed model Fed-MUnet is decentralized and other benchmarks are centralized.}}
\label{tab:result}
\begin{center}
\begin{tabular}{|c|c|c|c|c|c|c|c|}
\hline
\textbf{}&\multicolumn{7}{c|}{\textbf{Quantification}} \\
\hline
\textbf{Method} & \textbf{\textit{Metrics}}& \textbf{\textit{Dice score}}& \textbf{\textit{Jaccard index}}& \textbf{\textit{Hausdorff distance}} & \textbf{\textit{Sensitivity}} & \textbf{\textit{Precision}}& \textbf{\textit{Specificity}}\\
\hline

&ET&0.829±0.167&0.751±0.254&3.582±6.597&0.819±0.255&0.875±0.140&0.994±0.005\\
nnU-Net\cite{A25}&TC&0.875±0.136&0.813±0.247&5.219±10.210&0.856±0.243&0.917±0.122&0.996±0.006\\
&WT&0.919±0.060&0.858±0.113&5.974±13.016&0.899±0.113&0.939±0.062&0.996±0.003\\
\hline
&ET&0.834±0.185&0.748±0.206&6.915±13.715&0.842±0.199&0.862±0.168&0.992±0.004 \\
UNETR\cite{A26}&TC&0.847±0.206&0.776±0.228&9.609±15.458&0.846±0.214&0.890±0.166&0.995±0.004 \\
&WT&0.902±0.099&0.834±0.132&10.454±18.364&0.896±0.121&0.918±0.097&0.997±0.005 \\
\hline
\textbf{Fed-MUnet}&\textbf{ET}&\textbf{0.840±0.171}&\textbf{0.750±0.184}&\textbf{5.134±10.191}&\textbf{0.824±0.191}&\textbf{0.880±0.150}&\textbf{0.999±0.002}\\
\textbf{with}&\textbf{TC}&\textbf{0.878±0.144}&\textbf{0.805±0.178}&\textbf{10.347±17.100}&\textbf{0.875±0.154}&\textbf{0.908±0.129}&\textbf{0.998±0.005}\\
\textbf{$\boldsymbol{Noise} \boldsymbol{\sim} \boldsymbol{N(0, 10^{-2})}$}&\textbf{WT}&\textbf{0.905±0.094}&\textbf{0.837±0.128}&\textbf{11.383±17.110}&\textbf{0.895±0.121}&\textbf{0.925±0.075}&\textbf{0.995±0.006}\\
\hline
&ET&0.842±0.169&0.756±0.195&5.429±10.275&0.848±0.185&0.867±0.157&0.992±0.002\\
nnFormer\cite{A27}&TC&0.859±0.186&0.789±0.215&10.089±17.450&0.870±0.188&0.883±0.167&0.995±0.002\\
&WT&0.909±0.828&0.843±0.118&9.427±16.778&0.906±0.110&0.922±0.795&0.996±0.003
\\
\hline
&ET&0.843±0.181&0.760±0.204&5.060±11.903&0.839±0.202&0.882±0.153&0.994±0.002\\
MultiFormer\cite{A28}&TC&0.856±0.196&0.786±0.223&7.383±13.025&0.848±0.210&0.908±0.149&0.995±0.003\\
&WT&0.911±0.087&0.846±0.120&8.820±16.547&0.904±0.115&0.927±0.078&0.997±0.005\\
\hline
&ET&0.850±0.177&0.770±0.200&4.330±9.708&0.844±0.201&0.892±0.139&0.994±0.003\\
TransBTS\cite{A7}&TC&0.873±0.189&0.810±0.214&6.043±11.971&0.859±0.205&0.921±0.146&0.996±0.005\\
&WT&0.913±0.081&0.849±0.115&9.096±16.448&0.913±0.104&0.921±0.079&0.996±0.004\\
\hline
&ET&0.852±0.176&0.772±0.199&3.789±9.510&0.847±0.197&0.882±0.153&0.994±0.003\\
MultiCNN\cite{A29}&TC&0.873±0.191&0.811±0.216&5.122±10.265&0.861±0.202&0.916±0.124&0.994±0.005\\
&WT&0.915±0.082&0.852±0.115&7.420±13.502&0.902±0.110&0.937±0.066&0.997±0.007\\
\hline
&ET&0.857±0.167&0.777±0.192&3.519±6.246&0.860±0.181&0.883±0.154&0.994±0.002\\
TuningUNet\cite{A30}&TC&0.875±0.179&0.811±0.208&5.831±9.994&0.869±0.191&0.912±0.143&0.996±0.003\\
&WT&0.919±0.073&0.859±0.110&5.920±11.229&0.907±0.107&0.942±0.057&0.997±0.003\\
\hline
\textbf{Fed-MUnet}&\textbf{ET}&\textbf{0.860±0.164}&\textbf{0.778±0.176}&\textbf{3.870±9.777}&\textbf{0.859±0.181}&\textbf{0.882±0.144}&\textbf{0.998±0.002}\\
\textbf{with}&\textbf{TC}&\textbf{0.894±0.141}&\textbf{0.830±0.173}&\textbf{5.348±10.514}&\textbf{0.895±0.144}&\textbf{0.914±0.130}&\textbf{0.998±0.006}\\
\textbf{$\boldsymbol{Noise} \boldsymbol{\sim} \boldsymbol{N(0, 10^{-5})}$}&\textbf{WT}&\textbf{0.916±0.083}&\textbf{0.857±0.118}&\textbf{7.361±13.085}&\textbf{0.904±0.116}&\textbf{0.938±0.060}&\textbf{0.996±0.004}\\
\hline
\textbf{Fed-MUnet}&\textbf{ET}&\textbf{0.862±0.157}&\textbf{0.781±0.174}&\textbf{3.370±5.781}&\textbf{0.871±0.175}&\textbf{0.876±0.136}&\textbf{0.998±0.002}\\
\textbf{with}&\textbf{TC}&\textbf{0.897±0.140}&\textbf{0.834±0.171}&\textbf{4.961±9.658}&\textbf{0.892±0.146}&\textbf{0.923±0.117}&\textbf{0.998±0.004}\\
\textbf{no noise}&\textbf{WT}&\textbf{0.921±0.087}&\textbf{0.861±0.120}&\textbf{6.716±12.820}&\textbf{0.908±0.116}&\textbf{0.945±0.061}&\textbf{0.997±0.005}\\
\hline
\end{tabular}
\label{tab1}
\end{center}
\vspace{-1.5mm}
\end{table*}



\subsubsection{\textbf{Implementation details}}

The parameter $\alpha$ in \mbox{Dirichlet} distribution controls the sharpness of the distribution. If \(\alpha < 1\), the distribution will be highly unbalanced. However, the dataset is not sufficient to set \(\alpha < 1\) since some of the local dataset $X_K$ may be extremely small if $\alpha < 1 $, which makes it hard for the local model to learn. Therefore, in the experimental phase, we set \(\alpha = 2\) to make the dataset distribution slightly more evenly. 

In our implementation, all methods use the same training settings with mainstream experiments. The optimizer is SGD with a momentum of 0.95. The learning rate is $10^{-2}$ and the batch size is 32. In FL setting, we train for 30 global rounds to ensure that the model converges steadily and that the local update epoch is set to 3. In local rounds, each client K has a 50\% probability (q=0.5) of being selected to perform a local epoch. Our model is trained on an RTX 4090 GPU and implemented using PyTorch.

\subsubsection{\textbf{Evaluation metrics}}
Six metrics are used to assess the performance of the model in different aspects. Specifically, the Dice score, the harmonic mean of precision and recall, is used to assess the degree of overlap between segmentation results and ground truth labels. The Jaccard index calculates the ratio of the intersection size to the union size between the \mbox{segmentation} result and the ground truth label. The \mbox{Hausdorff} distance measures the greatest separation between two edge contours, providing a more sophisticated assessment of the segmented edges’ accuracy. Sensitivity is defined as the \mbox{percentage} of true positive cases (TP) overall true tumor regions (TP+FN). Precision measures the percentage of true positive cases (TP) over all regions predicted to be tumors (TP+FP). Specificity estimates the percentage of true negative cases (TN) among all non-tumor regions (TN+FP). Moreover, we evaluate the mean and standard deviation of the six metrics on enhancing tumor (ET), tumor core (TC), and whole tumor (WT) respectively. 
\vspace{-5pt}
\begin{figure}
\vspace{-3pt}
\vspace{-1.5mm}
\centering
\includegraphics[height=6.5cm,width=9cm]{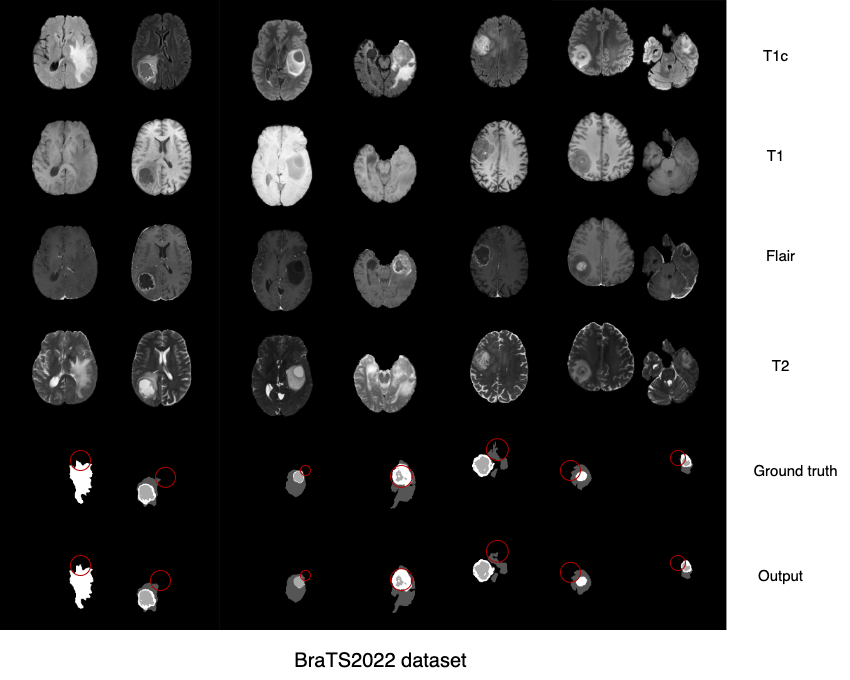}
\caption{\small{ Examples of Segmentation Results and Ground Truth. The area marked by the red circle is the difference between the true label and the inference result of the model with $\sigma = 10^{-5}$.}}
\label{3}
\vspace{-1mm}
\end{figure} 

\subsection{\textbf{Experimental Results}}

 Fig \ref{3} displays some segmentation results of our proposed model with $\sigma = 10^{-2}$, $\sigma = 10^{-5}$ and no noise, and other centralized model have the same experimental setting. The results of existing centralized models refer to the article \cite{A8}.

As shown in \textbf{Table \ref{tab:result}}, our model, Fed-MUnet with $\sigma\!=\!10^{-5}$, demonstrates substantial performance while ensuring data \mbox{privacy}. For the Dice Score, Fed-MUnet with noise achieves commendable values of 0.860 for Enhancing Tumor (ET), 0.894 for Tumor Core (TC), and 0.916 for Whole Tumor (WT) indicating a high degree of overlap with the ground truth. In the Jaccard Index, the model achieves results of 0.778 for ET, 0.830 for TC, and 0.853 for WT, further proving its segmentation accuracy. The Hausdorff Distance metrics demonstrates competitive performance with values of 3.870 for ET, 5.348 for TC, and 7.361 for WT. Sensitivity results, which are crucial for identifying true positive rates, are robust at 0.859 for ET, 0.895 for TC, and 0.904 for WT. Precision, reflecting the model’s ability to avoid false positives, is also high, with scores of 0.882 for ET, 0.914 for TC, and 0.919 for WT. Finally, the model achieves near-perfect specificity across all tumor regions, underscoring its reliability in distinguishing non-tumor areas. Although Fed-MUnet without noise showed slightly superior results across most metrics, the marginal performance difference compared to Fed-MUnet with noise highlights the latter’s robustness in protecting data privacy without significantly compromising accuracy. 

Furthermore, we evaluate the complexity of our \mbox{proposed} framework based on the total number of parameters, the \mbox{quantity} of floating-point operations (FLOPs), and the \mbox{average} \mbox{inference time}. As shown in \textbf{Table \ref{tab:complexity}}, our model achieves a \mbox{balanced} performance across these metrics while \mbox{significantly} improves segmentation accuracy. \mbox{Specifically}, Fed-MUnet contains 71.61 million parameters, which is \mbox{moderate} \mbox{compared} to other models such as MultiCNN with 177.31 \mbox{million} parameters and UNETR with 90.86 million parameters. The computational complexity of Fed-MUnet, as indicated by FLOPs, is 13.98 Giga, which is lower than models like nnU-Net (59.53 Giga) and \mbox{MultiCNN} (91.01 Giga), reflecting its \mbox{efficiency} in \mbox{computation}. \mbox{Furthermore}, Fed-MUnet’s \mbox{inference} time stands at 0.186 seconds, \mbox{demonstrating} its \mbox{competitive} speed relative to other models such as UNETR (0.208 seconds) and MultiCNN (0.239 \mbox{seconds}). Notably, our model is the Pareto optimal except for TransBTS and MultiFormer, but achieves higher performance and tackled privacy issue.

These results demonstrate that Fed-MUnet achieves an \mbox{optimal} balance between parameter count, computational \mbox{demand}, and inference time, making it a suitable modal for high-precision brain tumor segmentation in a federated learning framework.

\begin{table}[htbp]
\vspace{-5pt}
\caption{Evaluation of Model Complexity}
\label{tab:complexity}
\begin{center}
\begin{tabular}{|c|c|c|c|}
\hline
\textbf{}&\multicolumn{3}{c|}{\textbf{Metrics}} \\
\hline
\textbf{Method} & \textbf{\textit{Params(M)}}& \textbf{\textit{FLOPs(G)}}& \textbf{\textit{Inference time(s)}}\\
\hline
nnU-Net\cite{A25}&64.14&59.53&0.159\\
UNETR\cite{A26}&90.86&12.86&0.208\\
nnFormer\cite{A27}&89.29&12.41&0.198\\
MultiFormer\cite{A28}&89.97&12.72&0.122\\
TransBTS\cite{A7}&24.23&12.84&0.117\\
MultiCNN\cite{A29}&177.31&91.01&0.239\\
TuningUNet\cite{A30}&59.12&27.65&0.188\\
\textbf{Fed-MUnet}&\textbf{71.61}&\textbf{13.98}&\textbf{0.186}\\

\hline
\multicolumn{4}{l}{}
\end{tabular}
\label{tab1}
\end{center}
\end{table}

\vspace{-25pt}
\section{\textbf{Conclusion}}
 In this paper, we proposed Fed-MUnet for segmenting brain tumors from multi-modal MRI images. The proposed segmentation backbone M-Unet could be \mbox{easily} deployed in FL setting, addressing the privacy issue in \mbox{medical} big data. We conducted experiments using the publicly \mbox{available} BraTS2022 dataset to demonstrate the \mbox{effectiveness} of our algorithm. Our framework Fed-MUnet achieved the highest performance on brain tumor segmentation and \mbox{preserved} data privacy with few parameters and low model complexity.

In future work, we will investigate more sophisticated multi-modal fusion techniques to better secure data privacy without adversely affecting segmentation accuracy. In addition, the modal heterogeneity of MRI datasets mainly refers to the data difference and diversity caused by different imaging \mbox{parameters}, sequence types, equipment manufacturers and other factors in the MRI imaging process. We will explore advanced FL algorithm to modal heterogeneity in MRI.


\vspace{-5pt}
\begin{thebibliography}{00}
\bibitem{A30}
M. Futrega, M. Marcinkiewicz, and P. Ribalta, “Tuning U-Net for brain tumor segmentation,” in \textit{Brainlesion: Glioma, Multiple Sclerosis, Stroke and Traumatic Brain Injuries}, S. Bakas et al., Eds. Cham, Switzerland: Springer, 2023, pp. 162–173.


\bibitem{A17} Choudhury, Ahana Roy, Rami Vanguri, Sachin R. Jambawalikar and Piyush Kumar.(2018). “Segmentation of Brain Tumors Using DeepLabv3+.” in \textit{BrainLes@MICCAI}.


\bibitem{A7} W. Wang, C. Chen, M. Ding, H. Yu, S. Zha, and J. Li, “TransBTS: Multimodal brain tumor segmentation using transformer,” in \textit{Proc. 24th Int. Conf. Med. Image Comput. Comput.-Assist. Intervent.} Cham, Switzerland: Springer, 2021, pp. 109–119.

\bibitem{A8}
H. Liu, Z. Ni, D. Nie, D. Shen, J. Wang, and Z. Tang, “Multimodal brain tumor segmentation boosted by monomodal normal brain images,”  in\textit{IEEE Transactions on Image Processing}, vol. 33, pp. 1199–1210, Jan. 2024, doi: 10.1109/tip.2024.3359815.


\bibitem{A10}
 U.S. Department of Health and Human Services, "Health Insurance Portability and Accountability Act of 1996 (HIPAA)," Public Law 104-191, Aug. 1996. Available: https://www.hhs.gov/hipaa

\bibitem{A11}
European Parliament and Council of the European Union, "Regulation (EU) 2016/679 of the European Parliament and of the Council of 27 April 2016 on the protection of natural persons with regard to the processing of personal data and on the free movement of such data (General Data Protection Regulation)," \textit{Official Journal of the European Union}, L119, pp. 1-88, May 2016. 



\bibitem{f2}
B. Luo, W. Xiao, S. Wang, J. Huang, and L. Tassiulas, “Cost-Effective Federated Learning Design,” \textit{IEEE INFOCOM 2021 - IEEE Conference on Computer Communications}, Oct 2021, doi: 10.1109/infocom42981.2021.9488679.

\bibitem{f3}
B. Luo, W. Xiao, S. Wang, J. Huang, and L. Tassiulas, “Tackling System and Statistical Heterogeneity for Federated Learning with Adaptive Client Sampling,” \textit{IEEE INFOCOM 2022 - IEEE Conference on Computer Communications}, May 2022, doi: 10.1109/infocom48880.2022.9796935.


\bibitem{A12} Y. Zhao, Q. Liu, X. Liu, and K. He, "Medical Federated Model with Mixture of Personalized and Sharing Components," in \textit{Proceedings of the IEEE}, vol. 10, no. 5, pp. 1-10, 2022.


\bibitem{A14} Li, W., Milletarì, F., Xu, D., Rieke, N., Hancox, J., Zhu, W., Baust, M., Cheng, Y., Ourselin, S., Cardoso, M. J., \& Feng, A. (2019). Privacy-Preserving Federated Brain Tumour Segmentation. In Suk, H.I., Liu, M., Yan, P., Lian, C. (Eds.) Machine Learning in Medical Imaging. MLMI 2019. Lecture Notes in Computer Science, vol 11861. Springer, Cham. 

\bibitem{A15} Ullah, Faizan, Muhammad Nadeem, Mohammad Abrar, Farhan Amin, Abdu Salam, and Salabat Khan, “Enhancing Brain Tumor Segmentation Accuracy through Scalable Federated Learning with Advanced Data Privacy and Security Measures,” Mathematics, vol. 11, no. 19, p. 4189, Oct. 2023. 

\bibitem{A9} O. Ronneberger, P. Fischer, and T. Brox, "U-Net: Convolutional Networks for Biomedical Image Segmentation," in \textit{Proceedings of the International Conference on Medical Image Computing and Computer-Assisted Intervention (MICCAI)}, Munich, Germany, 2015, pp. 234-241.


\bibitem{A19} R. Geyer, T. Klein, and M. Nabi, ``Differentially Private Federated Learning: A Client Level Perspective,'' \textit{arXiv:1712.07557}, 2017.

\bibitem{A20}
Drozdzal, M., Vorontsov, E., Chartrand, G., Kadoury, S., \& Pal, C. (2016). "The Importance of Skip Connections in Biomedical Image Segmentation". In \textit{Deep Learning and Data Labelling for Medical Applications} (pp. 179-187). Springer, Cham.


\bibitem{A22}
H. Tzu-Ming Harry Hsu and M. Brown.(2019). "Measuring the Effects of Non-Identical Data Distribution for Federated Visual Classification," arXiv, Sep. [Online]. Available: https://arxiv.org/abs/1909.06335

\bibitem{A23}
S. Guo, X. Yang, J. Feng, Y. Ding, W. Wang, Y. Feng, and Q. Liao, "FedGR: Federated Learning with Gravitation Regulation for Double Imbalance Distribution," in \textit{Database Systems for Advanced Applications - 28th International Conference, DASFAA}, April 17-20, 2023, pp. 1-16..


\bibitem{A25}
F. Isensee, P. F. Jaeger, S. A. A. Kohl, J. Petersen, and K. H. Maier-Hein, “nnU-Net: A self-configuring method for deep learning-based biomedi- cal image segmentation,” Nature Methods, vol. 18, no. 2, pp. 203–211, Feb. 2021.

\bibitem{A26}
A. Hatamizadeh et al., “UNETR: Transformers for 3D medical image segmentation,” in \textit{Proc. IEEE/CVF Winter Conf. Appl. Comput. Vis. (WACV)}, Jan. 2022, pp. 574–584.

\bibitem{A27}
H.-Y. Zhou, J. Guo, Y. Zhang, L. Yu, L. Wang, and Y. Yu,(2021), “NnFormer: Interleaved transformer for volumetric segmentation,” ,\textit{arXiv:2109.03201}.

\bibitem{A28}
J. Cho and J. Park, “Multi-modal transformer for brain tumor segmenta-tion,” in Brainlesion: Glioma, Multiple Sclerosis, Stroke and Traumatic Brain Injuries, S. Bakas et al., Eds. Cham, Switzerland: Springer, 2023, pp. 138–148.

\bibitem{A29}
R. A. Zeineldin, M. E. Karar, O. Burgert, and F. Mathis-Ullrich, “Multimodal CNN networks for brain tumor segmentation in MRI: A brats 2022 challenge solution,” in \textit{Brainlesion: Glioma, Multiple Sclerosis, Stroke and Traumatic Brain Injuries}, S. Bakas et al., Eds. Cham, Switzerland: Springer, 2023, pp. 127–137.

\end{thebibliography}
\end{document}